\documentclass{article}
\usepackage{amsfonts}
\usepackage{adjustbox}
\usepackage{amssymb}
\usepackage{spconf,amsmath,graphicx}
\usepackage{multirow}
\usepackage{subfigure}
\usepackage{algorithm}
\usepackage{algpseudocode}
\usepackage{booktabs}
\usepackage{url}
\title{Federated Learning of Large ASR Models in the Real World}
\name{Yonghui Xiao \qquad Yuxin Ding \qquad Changwan Ryu \qquad Petr Zadrazil \qquad Fran\c{c}oise Beaufays}
\address{
  Google LLC, Mountain View, CA, U.S.A }
\begin{document}
%\ninept
%
\maketitle
\begin{abstract}
Federated learning (FL) has shown promising results on training machine learning models with privacy preservation. However, for large models with over 100 million parameters, the training resource requirement becomes an obstacle for FL because common devices do not have enough memory and computation power to finish the FL tasks. Although efficient training methods have been proposed, it is still a challenge to train the large models like Conformer based ASR. This paper presents a systematic solution to train the full-size ASR models of 130M parameters with FL. To our knowledge, this is the first real-world FL application of the Conformer model, which is also the largest model ever trained with FL so far. And this is the first paper showing FL can improve the ASR model quality with a set of proposed methods to refine the quality of data and labels of clients. We demonstrate both the training efficiency and the model quality improvement in real-world experiments.
\end{abstract}
\begin{keywords}
federated learning, speech recognition
\end{keywords}
\section{Introduction}
\label{sec:intro}
Federated learning (FL) has shown promising results on training machine learning (ML) models with privacy preservation \cite{DBLP:journals/corr/McMahanMRA16,guliani2021training}. Because FL has access to the on-device data which is not available on centralized server-side training, it's specially good at learning on-device related patterns, e.g. whether users have feedback to the on-device apps. Moreover, FL can also be combined with centralized server training under a joint training framework \cite{DBLP:journals/corr/abs-2111-12150} to  mitigate distribution shift of FL and further improve the model quality.

With the above advantages, one of the drawbacks of FL is that the models have to be trained on users' devices where only limited resources such as memory and computation power are available. The problem becomes worse given that recent models are getting larger and larger such as large language models (LLM) ChatGPT \cite{liu2023summary} and PaLM \cite{anil2023palm}. For end-to-end automatic speech recognition (ASR) models \cite{wang2019overview,streaming_conformer}, the performance is also subject to the model size. Specifically, the Conformer-based ASR model \cite{gulati2020conformer,streaming_conformer, recognizing_long_form} usually requires 120$\sim$150 million parameters to achieve the desired recognition quality. Models of this size requires several GB of training memory, e.g. \cite{sim2019personalization}, hence it is a big challenge for FL.

Efficient FL methods have been proposed to relieve the resource burden on FL devices. There are generally two categories of the related works. First, from the model perspective new training algorithms are proposed, including pruning technique \cite{DBLP:conf/interspeech/DingWZSHDBWPLHM22} and dropout method \cite{JMLR:v15:dropout} to reduce the model size, gradient checkpointing \cite{DBLP:journals/corr/ChenXZG16} to recompute the gradients in backward propagation and quantization method \cite{ding20224,zhen2023sub} to reduce the variables precision. Second, FL related algorithms are designed, including federated dropout\cite{guliani2022enabling} to train smaller models on clients, federated pruning\cite{lin2022federated} to reduce the overall model size, online model compression (OMC) \cite{yang2022online} to quantize the model to lower precision and partial variables training \cite{yang2021partial} to only compute partial gradients to save the memory usage.
However, the above works only focus on one aspect of the system and it's unknown if the integration of different approach would enable the FL of ASR models.

\begin{figure}[t]
  \centering
  \includegraphics[width=\linewidth]{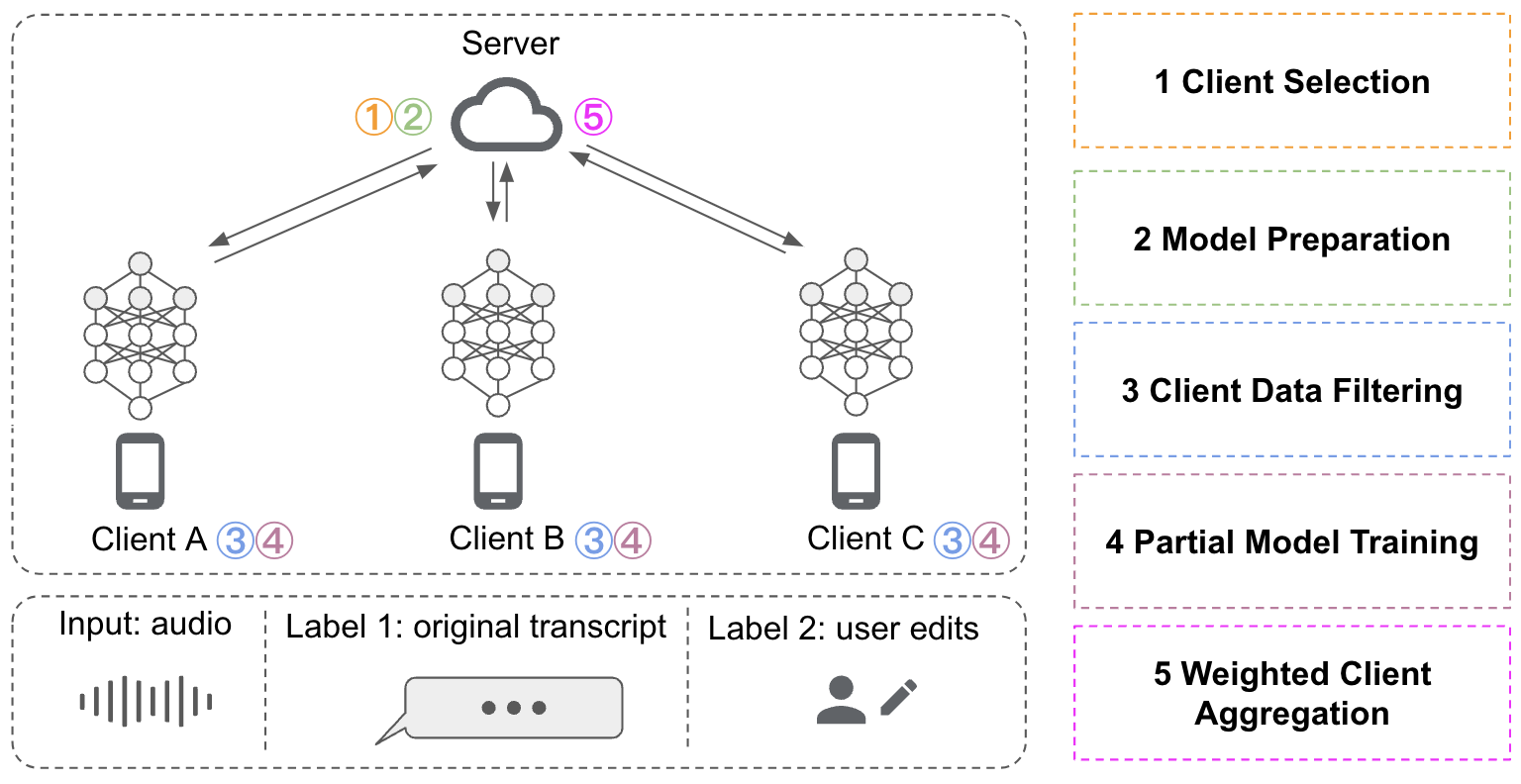}
  \caption{The overview of the FL system. 1$\sim$5 are the FL steps in a round. There are 3 types of data on clients: the input audio data, the original transcript from the incumbent ASR and the final transcript based on user edits.}
  \label{fig:learning_process}
\end{figure}

In this paper we study the FL of Conformer based ASR model with 130M parameters, which is the largest model for FL so far. Our work potentially paves the road to train other large models like LLMs in the future. We first build different methods into one system and show how the consolidated system works together. Then we study the problem of how to improve the model quality with FL. Because FL is good at learning the on-device usage patterns, we design the FL algorithm as follows. For each user, there is an incumbent ASR on the device to generate the transcript from user audio. We observed that users might edit the original transcripts to correct the errors. For example, if a user said ``covid" and the incumbent ASR outputs ``covert", then users may change the output to ``covid" again. Therefore, we utilize user-correction actions on devices to improve the model performance on the ”corrected” words. Figure \ref{fig:learning_process} highlights our approach. At the beginning of a FL round, the server selects the clients that have the correction actions. Then the server sends a ``processed" (e.g. quantized/pruning/reduced) model to clients. Clients runs a data filtering method to only use the ``correction" data as training examples, and partially train the model under the resource constraint. Then we design a weighted client aggregation (WCA) algorithm to update the trained model on server. Our contributions are summarized as follows.
\begin{itemize}
  \vspace{-2mm}
  \item To our knowledge, this is the first real-world FL application that successfully trains the production-grade ASR model of 130M parameters. We explain the FL system in Section \ref{sec-efficiency}.
  \vspace{-2mm}
  \item This is also the first paper that shows the ASR quality can be improved by FL. We propose the WCA algorithm to refine the data and label quality of clients based on the user-corrections, described in Section \ref{sec-quality}.
  \vspace{-2mm}
  \item We conducted real-world FL experiments to demonstrate the performance of our system in Section \ref{sec-exp}. We report that the training efficiency is greatly increased, which is measured by (1) the memory usage and (2) transportation size between server and clients, and the WER of FL models is effectively boosted.
\end{itemize}

\section{Training Efficiency}
\label{sec-efficiency}
In this section we describe how the FL system is built to train the ASR models. The bottleneck of the FL system consists of two constraints: (1) the on-device peak memory usage and (2) the transportation size between server and clients. The two constraints are also positively correlated as they can be optimized together.

First we combine the existing algorithms together in our system as a baseline, meaning that all the following methods are applied by default unless explained further. We enable the gradient checkpointing method \cite{DBLP:journals/corr/ChenXZG16} in the Conformer model to reduce the memory usage. Moreover, we use the FedSGD \cite{DBLP:journals/corr/McMahanMRA16} algorithm to only take one batch of data as we observed that more examples lead to more memory usage with on-device CPU training \cite{sim2019personalization}. To further reduce the memory consumption, we set a small batch size like 2. With this setting, it means the convergence speed might be slower compared to large batch size and FedAVG \cite{DBLP:journals/corr/McMahanMRA16}. But it can be compensated by more FL rounds and large report goal (the number of clients participating in one FL round). Between server and clients, transportation compression methods \cite{tff_http} are applied to reduce the network load. Next we add two methods on the baseline system including OMC and partial model training.

\textbf{OMC.} We build the OMC \cite{yang2022online} method in our system to reduce both the model size and the memory usage. 
% We implement the float16 OMC method because the precision loss of float16 is relatively small. Lower-precision is also applicable if the trade-off of slower convergence and quality degradation is acceptable. 
Specifically, the OMC method is the step $2$ in Figure \ref{fig:learning_process}. Before sending the model to clients, the server quantizes the variables to low-bit precision \cite{tf_float16}. To balance the quality degradation and the training efficiency, we quantize the matrices variables to float16 and keep other variables in the original float32 format as the model quality is more sensitive to the biases and activations. Then the server sends the quantized models to clients and clients compute gradients with the same precision of the variables. In this way, both the download and upload sizes are reduced with the float16 format. The memory usage is also reduced because variable storage memory of float16 variables is smaller compared to float32 while the gradient computation is bounded by gradient checkpointing \cite{DBLP:journals/corr/ChenXZG16}. The results are reported in our experiments.

\textbf{Partial model training.} We build an updated partial model training \cite{yang2021partial} corresponding to the step 4 in Figure \ref{fig:learning_process} to reduce the memory usage and the upload size. It sets a subset of trainable variables and non-trainable variables in the model. In this way only a subset of gradients, i.e. for the trainable variables, needs to be computed and uploaded. And the non-trainable variables stay frozen during the training. Moreover, we freeze consecutive bottom encoder layers and only set the decoder and top encoder layers as trainable. When combining it with OMC, we observed that partial model training converges slower. To boost the convergence speed, we de-quantize the trainable variables to float32 again. To summarize, float32 variables consist of (1) all trainable variables from the decoder and top encoder layers and (2) the activations in non-trainable variables from the bottom encoder layers. And float16 includes matrices in non-trainable variables from the bottom encoder layers.
The performance is then shown in our experiments.

% Federated dropout \cite{guliani2022enabling} and federated pruning \cite{lin2022federated} are also applicable but showed small improvements on top of our system. We will show the performance of federated dropout in our experiments and skip the discussion of these two methods.
\section{Model Quality}
\label{sec-quality}
We explain our algorithm to boost the model quality of FL. The high level idea is to utilize the user-correction actions to refine the quality of data and labels with weighted client aggregation in FL.

\textbf{Client selection.} To adopt the user-correction data, FL server selects the clients containing the corrections at the step 1 in Figure \ref{fig:learning_process}. At the beginning of an FL round, the server sends all clients an eligibility test designed to check if a client has the user-correction data. If a client passes the eligibility test, it will continue to participate the FL round. Otherwise, the client will drop out from the FL round. The server will keep sending the eligibility test until enough clients are collected to reach the expected report goal. In this way, all participating clients will have the correction data.

\textbf{Data filtering on devices.} At the step 3 in Figure \ref{fig:learning_process} when a client receives the model, the client needs to filter the data first. The purpose of the data filtering is to only take the user-correction data in the FL training. If the client batch size larger than 1, to make sure a client has enough data to form a batch of data after the filtering, we design the eligibility test to check if a client has enough ($>=$ the batch size) user-corrections. Another way is to duplicate the existing data to form a batch, which will change the training data distribution considered in the following WCA algorithm.

Another benefit of client selection and data filtering is to eliminate the ``incorrect" corrections, e.g. a user said ``covid" but got ``covert" as transcript, then the user changed the transcript to ``covertcovid" incorrectly. Because the true data is inaccessible in FL, such ``incorrect" corrections also participate in the FL training and pollute the training data. Therefore, we need to filter out such ``incorrected" examples. To do so, we use heuristics to estimate and quantify the quality of a correction in the eligibility test, e.g. the word length difference before and after a user edit should be smaller than a threshold. If the quality of correction is low by the hueristic, we eliminate the example.

\textbf{Weighted Clients Aggregation.} At the step 5 in Figure \ref{fig:learning_process}, the server aggregates all the client uploads, i.e. the gradients from FedSGD computation, together to update the server model. At this time, we propose a WCA algorithm to compute the server model update. The motivation of WCA is to align the distribution of training data to the target distribution to boost the training quality. In particular, our target distribution is based on the list of corrected words denoted by $\mathbb{W}$, i.e. a special distribution containing the incumbent model errors.

There are usually two aggregation methods in FL: (1) the simple averaging as $\Sigma_{i=1}^n{G_i}/n$ where $G_i$ is the model deltas of $n$ clients; and (2) the $\#$example based aggregation as $\Sigma_{i=1}^n{G_i}{E_i}/\Sigma_{i=1}^n{E_i}$ where $E_i$ is the number of participating examples of the each client in FedAVG. However, these aggregation methods have not considered the quality of the clients data. Thus the model quality may be degraded due to unexpected data as discussed before. To fix this problem, we propose a WCA algorithm as $\Sigma_{i=1}^n{G_i}{w_i}/\Sigma_{i=1}^n{w_i}$ where $w_i$ is the designed weights of $n$ clients as Algorithm \ref{alg:wca}.

\begin{algorithm}[h]
\caption{\emph{Weighted client aggregation}. $G$ is the gradient computed from an example. $w_{ij}$ is the designed weight of client $i$ example $j$.}
\label{alg:wca}
\begin{algorithmic}[1]

\For{each round $r$ = 1,2,...}
    \State{Server selects participating clients.}
    \State{Server sends prepared models to selected clients.}
    \For{each client $i \in n$ \textbf{in parallel}}
    \State{Filter examples to form a batch.}
    \For{each example $E_j$ in the batch}
        \State{Compute $w_{ij}$ for example $E_j$}
        \label{alg-line-weight}
    \EndFor
    \State{$G_i\gets \Sigma_{j}G_{ij}w_{ij}$}
    \State{$w_i \gets \Sigma_{j}w_{ij}$}
    \EndFor
    \State{Update the server model by $\Sigma_{i=1}^n{G_i}{w_i}/\Sigma_{i=1}^n{w_i}$}
\EndFor
\end{algorithmic}
\end{algorithm}

The key of WCA is how to design the weights $w_i$. 
Because user-correction pattern may be different from the training data, we need to make the best of the corrections.
For example, if the correction from ``pie torch" to ``pytorch" is rare, we need to assign higher weight to it. Otherwise the gradient contribution of the example will be submerged in the aggregated gradients and vanish in the learned model.
To this end, we propose two methods (1) {frequency based weights} and (2) {frequency and accuracy based weights}. Given the set of corrected words $\mathbb{W}$, the \textbf{frequency based weights} compute the frequency of each word in $\mathbb{W}$ among all clients. Such frequency can be derived by computing the differentially private histogram \cite{dpcube} of words on the client pool, i.e. $\mathbf{freq}_w$ is the differentially private frequency of word $w$. Then for each word $w$ in the example $j$ containing the corrected transcript of client $i$ at line \ref{alg-line-weight} in Algorithm \ref{alg:wca}, the $w_{ij}$ can be computed as follows.
\begin{equation}
\vspace{-2mm}
\label{eqn-frequency}
w_{ij} = \Sigma \cfrac{1}{\mathbf{freq}_w}, {w} \in \mathbf{correction}_j
\end{equation} In this way, we have higher weights for rare words and less weights for frequent words in $\mathbb{W}$. Based on this, the \textbf{frequency and accuracy based weights} also incorporate the word accuracy of the incumbent ASR model that generates the transcripts in the first place. The accuracy $\mathbf{acc}_w \in [0, 1]$ denotes how well the incumbent model recognizes a word $w$. The higher accuracy mean the word is recognized well and lower accuracy means the word is recognized poorly. Then the $w_{ij}$ can be computed as Equation \ref{eqn-freq-acc}.
\begin{equation}
\vspace{-2mm}
\label{eqn-freq-acc}
w_{ij} = \Sigma \cfrac{1-\mathbf{acc}_w}{\mathbf{freq}_w}, {w} \in \mathbf{correction}_j
\end{equation}

% \begin{equation}
% \vspace{-2mm}
% \label{eqn-freq-acc}
% w_{ij} = \Sigma \cfrac{1-\mathbf{acc}_w}{\mathbf{freq}_w}{\mathbf{Pr}(w)}, {w} \in \mathbf{correction}_j
% \end{equation}

\section{Experimental Results}
\label{sec-exp}
\subsection{Experiment Settings}
\textbf{Training settings.} We prepare a centrally pre-trained model at the server to warm start the FL training. The initial model was trained on a multi-domain datasets collected from domains of YouTube, farfield and telephony etc \cite{misra21_interspeech, recognizing_long_form}. All datasets are anonymized and our work abides by Google AI Principles \cite{google}. Our batch size is 2 and report goal is 128. All experiments were conducted on the real-world FL with users' smartphones including Google Pixel phones.

\textbf{Model Architecture.} Because our objective is to train the large ASR models, we chose the production-grade Conformer \cite{streaming_conformer, recognizing_long_form} model that has about 130M parameters. The model consists of a causal encoder for streaming case and another non-causal encoder for non-streaming case. We only train the causal encoder and the decoder for the streaming cases, although our method can be easily extended to the non-causal encoders. 

\textbf{Metrics.} To evaluate the FL training efficiency, we measure the metrics of transportation size between server and client, and the averaged clients peak memory usage. For the model quality, we use two WERs: (1) the ``general WER" refers to the WER on all evaluation datasets; and (2) ``target WER" refers to the WER on the utterances containing only the corrected data in $\mathbb{W}$. Because the objective is to improve the quality on the correction dataset, we mainly focus on the ``target WER" while maintaining the general WER at the same level. The baseline WERs from the pre-trained model is ``general WER" $\textbf{4.4}$ and ``target WER" $\textbf{17.5}$.
% \begin{table}[h]
% %   \small
%   \caption{WER baseline}
%   \label{tab:wer_baseline}
%   \centering
%   \scalebox{1.0}{
%   \begin{tabular}{c|c}
%     \toprule
%     \textbf{General WER}& 
%                              \textbf{Target WER} \\
%     \midrule
%     4.4 & 17.5   \\
%   \bottomrule
% \end{tabular}
% }
% % \vspace{-5mm}
% \end{table}

\subsection{Training Efficiency}
We first report the training efficiency metrics, which essentially are the bottleneck for training large ASR models in FL.

\textbf{OMC.} To evaluate the benefit of OMC method, we performed experiments on a 6-layer encoder Conformer model (under a 250MB download size constraint) to compare the metrics with and without OMC. Table \ref{tab:efficiency_omc} shows the result. The OMC method reduces the download and upload size by 40MB and 25MB. The peak memory usage is also reduced by about 150MB. Note that the transportation compression methods \cite{tff_http} are applied by default, and hence the transportation size is smaller than parameter memory size.
\begin{table}[h]
\vspace{-5mm}
%   \small
  \caption{Training efficiency vs OMC}
  \label{tab:efficiency_omc}
  \centering
  \scalebox{1.0}{
  \begin{tabular}{c|ccc}
    \toprule
    \textbf{Training setup}& 
                             \textbf{Download} & \textbf{Upload} & \textbf{Memory} \\
    \midrule
    No OMC & 131MB & 31MB & 965MB  \\
    with OMC & 91MB & 6MB & 819MB \\
  \bottomrule
\end{tabular}
}
% \vspace{-5mm}
\end{table}
% \textbf{Federated dropout.} Next we show the benefit of federated dropout in Table \ref{tab:efficiency_fd}.
% \begin{table}[h]
% %   \small
%   \caption{Training efficiency vs federated dropout}
%   \label{tab:efficiency_fd}
%   \centering
%   \scalebox{1.0}{
%   \begin{tabular}{c|ccc}
%     \toprule
%     \textbf{Training setup}& 
%                              \textbf{Download} & \textbf{Upload} & \textbf{Memory} \\
%     \midrule
%     No FD & 272MB & 182MB & 1.34GB  \\
%     FD 0.2 & 236MB & 164MB & 1.0GB \\
%     FD 0.5 & 185MB & 126MB & 870MB \\
%   \bottomrule
% \end{tabular}
% }
% % \vspace{-5mm}
% \end{table}

\textbf{Partial model training.} Next we report the evaluation of partial model training in Table \ref{tab:efficiency_pmt}. The upload size is reduced because only the gradients of trainable variables are uploaded. The peak memory usage is also reduced because the non-trainable layers are frozen without gradient computation. Because float16-OMC method is applied by default, partial model training actually increases the download size because the trainable parameters are extended to float32 precision. 
\begin{table}[h]
\vspace{-5mm}
%   \small
  \caption{Training efficiency vs partial model training}
  \label{tab:efficiency_pmt}
  \centering
  \scalebox{1.0}{
  \begin{tabular}{c|ccc}
    \toprule
    \textbf{Training setup}& 
                             \textbf{Download} & \textbf{Upload} & \textbf{Memory} \\
    \midrule
    Full model & 200MB & 72MB & 1.34GB \\
    Dec + 1L Enc & 231MB & 29MB & 727MB \\
    Decoder only & 247MB & 16MB & 677MB \\
    %Dec + 2L Enc & 298MB & 77MB & 755MB \\
  \bottomrule
\end{tabular}
}
% \vspace{-5mm}
\end{table}
\subsection{Model Quality}
We report the quality of the trained model in this section. The model was trained with both OMC and partial model training methods. Specifically we train the top-1 encoder layer and decoder with the bottom encoder layers being frozen as non-trainable variables. The ablation studies w.r.t. OMC and partial model training were also conducted with no significant findings, i.e. float16-OMC is close to the float32 precision; and more trainable variables improve the model quality. Hence we skip the report of ablation studies.

\textbf{WER improvement.} The results are summarized in Table \ref{tab:wer_improvement} where each row adds a new method to the above row. The initial FL had general WER 4.4 and target WER 17.2. When the client selection method was added, the quality is not improved because clients might use unrelated examples. Thus we add the data filtering method to specify the training examples, and achieved general WER 4.4 and target WER 16.9. Next we added the two WCA methods, and the frequency and accuracy based WCA obtained the best result of general WER 4.4 and target WER 14.9.
\begin{table}[h]
\vspace{-5mm}
%   \small
  \caption{WER of trained models}
  \label{tab:wer_improvement}
  \centering
  \scalebox{1.0}{
  \begin{tabular}{c|cc}
    \toprule
    &\textbf{General WER}& 
                             \textbf{Target WER} \\
    \midrule
    initial FL &4.4 & 17.2   \\
    + client selection & 4.5 & 17.2 \\
    + data filtering & 4.4 & 16.9\\
    + frequency WCA & 4.4 & 16.4\\
    + freq-accuracy WCA & 4.4 & 14.9\\
  \bottomrule
\end{tabular}
}
% \vspace{-5mm}
\end{table}

\begin{figure}[htpb]
\vspace{-5mm}
\centering
\subfigure[General WER along FL rounds]{
    \label{fig:wer_general}
    \includegraphics[width=0.47\linewidth]{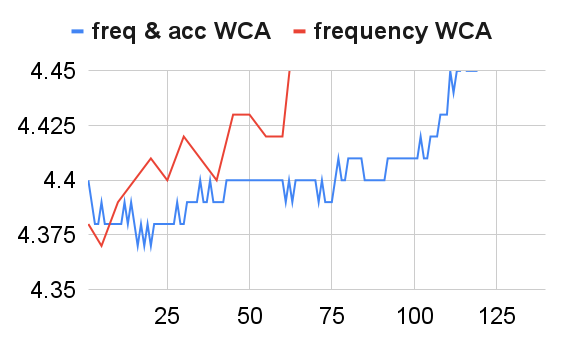}}
\subfigure[Target WER along FL rounds]{
    \label{fig:wer_target}
    \includegraphics[width=0.47\linewidth]{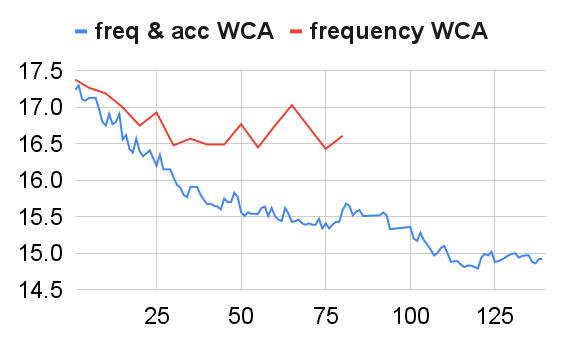}}
\vspace{-2mm}
\caption{WER trade-off between general WER and target wER.}
\label{fig:wers}
\end{figure}
\textbf{WER trade-off.} Because our objective is to improve the target WER, we need to consider the WER trade-off between the general WER and target WER. Figure \ref{fig:wers} shows the convergence curves of the two WCA methods. We can see that in Figure \ref{fig:wer_general} the general WER started to deteriorate after a FL round while the target WER keeps getting better in Figure \ref{fig:wer_target}. To balance the two WERs, we keep the general WER under the same level of 4.4 and take the corresponding target WER. Advanced trade-off can be designed in future works to further improve the performance.

% \textbf{WER vs partial model training.} We show the model quality w.r.t. the trainable parameters in Figure \ref{fig:wers_pmt}. We configured two settings: (1) partial model training with decoder-only as trainable variables; (2) partial model training with decoder and the top-1 encoder. We can see that with more trainable variables the general WER degrades quicker whereas the target WER gets better. This is the consequence of the FL setting that we only trained on the targeted dataset on selected clients.
% \begin{figure}[htpb]
% \centering
% \subfigure[decoder-only]{
%     \label{fig:wer_general_pmt}
%     \includegraphics[width=0.47\linewidth]{figures/general_wer_pmt.png}}
% \subfigure[decoder and top 1L encoder]{
%     \label{fig:wer_target_pmt}
%     \includegraphics[width=0.47\linewidth]{figures/target_wer_pmt.png}}
% \caption{WER vs partial model training.}
% \label{fig:wers_pmt}
% \end{figure}

\section{Conclusions}
In this paper we reported the first real-world FL application to train the Conformer model of about 130 million parameters. And we proposed new algorithms to improve the FL model quality by utilizing the user corrections on devices. At last we demonstrated the performance of the FL system in real-world applications to verify that both the training efficiency and the model quality were improved.

\small
\bibliographystyle{IEEEbib}
\bibliography{refs}

\end{document}